\documentclass[conference,fleqn]{setup/IEEEtran}
\usepackage{multirow}
\usepackage{cite}

\usepackage{amsmath,amssymb,amsfonts}
\usepackage{algorithmic}
\usepackage{graphicx}
\usepackage{textcomp}
\usepackage{xcolor}

\usepackage[hyphens]{url}
\usepackage{todonotes} 

\usepackage{verbatim}
\usepackage{xcolor}
\usepackage[font=normalsize]{caption}
\begin{document}
    \title{FinLlama: Financial Sentiment Classification for Algorithmic Trading Applications}
    \author{\IEEEauthorblockN{Thanos Konstantinidis\(^{1}\), Giorgos Iacovides\(^{1,2}\), Mingxue Xu\(^{1}\), Tony G. Constantinides\(^{1}\), Danilo Mandic\(^{1}\)}
    \IEEEauthorblockA{\(^{1}\)Imperial College London, \(^{2}\)MIT \\
    \{a.konstantinidis16, giorgos.iacovides20, m.xu21, a.constantinides, d.mandic\}@imperial.ac.uk}}

    \maketitle
    \thispagestyle{plain}
    \pagestyle{plain}
    \begin{abstract}
    There are multiple sources of financial news online which influence market movements and trader's decisions. This highlights the need for accurate sentiment analysis, in addition to having appropriate algorithmic trading techniques, to arrive at better informed trading decisions. Standard lexicon based sentiment approaches have demonstrated their power in aiding financial decisions. However, they are known to suffer from issues related to context sensitivity and word ordering.  Large Language Models (LLMs) can also be used in this context, but they are not finance-specific and tend to require significant computational resources. To facilitate a finance specific LLM framework, we introduce a novel approach based on  the Llama 2 7B foundational model, in order to benefit from its generative nature and comprehensive language manipulation. This is achieved by fine-tuning the Llama2 7B model on a small portion of supervised financial sentiment analysis data, so as to jointly handle the complexities of financial lexicon and context, and further equipping it with a neural network based decision mechanism. Such a generator-classifier scheme, referred to as FinLlama, is trained not only to classify the sentiment valence but also quantify its strength, thus offering traders a nuanced insight into financial news articles. Complementing this, the implementation of parameter-efficient fine-tuning through LoRA optimises trainable parameters, thus minimising computational and memory requirements, without sacrificing accuracy. Simulation results demonstrate the ability of the proposed FinLlama to provide a framework for enhanced portfolio management decisions and increased market returns. These results underpin the ability of FinLlama to construct high-return portfolios which exhibit enhanced resilience, even during volatile periods and unpredictable market events.
    
\end{abstract}
    \textit{\textbf{Index Terms}} - Large language models, finance, sentiment analysis, algorithmic trading, parameter-efficient fine-tuning
    
    \section{Introduction}
    The ever increasing prominence of algorithmic trading in quantitative finance has necessitated the need for reliable and actionable AI-aided intelligence from vast streams of data with multiple modalities. Of particular interest is generative AI, owing to its ability to distill insights from non-numerical sources such as news, earnings calls, financial reports, and other textual sources. In this context, sentiment analysis promises to bridge the gap between market movements caused by geopolitical and socioeconomic events, human actions, and quantitative trading. \par 
    Sentiment analysis rests upon the quantification of  opinions present in unlabeled textual data, and aims to categorize whether the overall viewpoint is positive, negative, or neutral. When applied to large-scale information sources, this may yield an understanding of the overall direction of macroscopic trends, a task which is both challenging and time-consuming for human analysts. Importantly, the sentiment contained in on-line textual sources can drive market movements; such information harbours intrinsic advantages and a competitive edge to those equipped with the tools to harness it. \par 
    Despite conceptual benefits, the heterogeneous, nuanced, and vast nature of financial text presents unique challenges when it comes to extracting sentiment in a manner that is both accurate and actionable. For example, the words `bull’ and `bear’ are neutral in the general vocabulary, but in financial markets, their respective connotations are strictly positive or negative \cite{lexicon_approaches}. This highlights the need for context-aware sentiment extraction, and underpins the complexities of natural language processing (NLP) in financial applications. \par 
    To adress these issues, we consider a two-fold fundamental question:
    \begin{itemize}
        \item Can large language models (LLMs), which have revolutionized manifold areas of NLP, be specifically tailored for sentiment analysis in the finance domain, particularly for algorithmic trading?
        \item Can this tailoring be achieved in a way which does not require the vast computational resources, typically associated with NLP models, thus making the approach accessible to  a broader audience equipped with standard computational resources?
    \end{itemize}.  \par 
    Our proposed \textit{FinLlama} is one such solution, which is obtained by fine-tuning a pre-trained LLM (namely Llama 2 7B \cite{touvron2023llama}) on specialised, labelled and publicly available financial news datasets. The ultimate goal of FinLlama is to enhance the performance of financial sentiment analysis, whilst leveraging on parameter-efficient fine-tuning (PEFT) and 8-bit quantization, through LoRA \cite{hu2021lora}, to minimise resource utilisation. \par 
    The main contributions of this work are:
    \begin{itemize}
        \item \textbf{Targeted fine-tuning}: Rather than utilising one over-arching model for diverse financial tasks, our approach capitalizes on the foundational pre-trained Llama 2 model, whereby fine-tuning is performed specifically for the purpose of sentiment classification through a SoftMax classification layer at its output.
        \item \textbf{Efficient resource utilization}: Our approach ensures that even standard computational resources, with no high-end GPUs, can be employed. By virtue of the pre-trained Llama 2 model and through targeted parameter-efficient fine-tuning, we dramatically reduce computational demands compared to the existing methods, thus bridging the gap between academic benchmarks and practical utility.
        \item \textbf{Benchmarking and real-world application}: The success of fine-tuned LLMs for finance has also highlighted that the domain of portfolio construction has not yet been adequately addressed. To this end, we integrate the extracted sentiment signals by FinLlama into a long-short portfolio, which allows us to obtain finance-specific real-world metrics including cumulative returns and the Sharpe ratio. 
    \end{itemize}

    \section{Related Work}
\begin{figure*}[htbp] 
\centering 
\includegraphics[width=\textwidth]{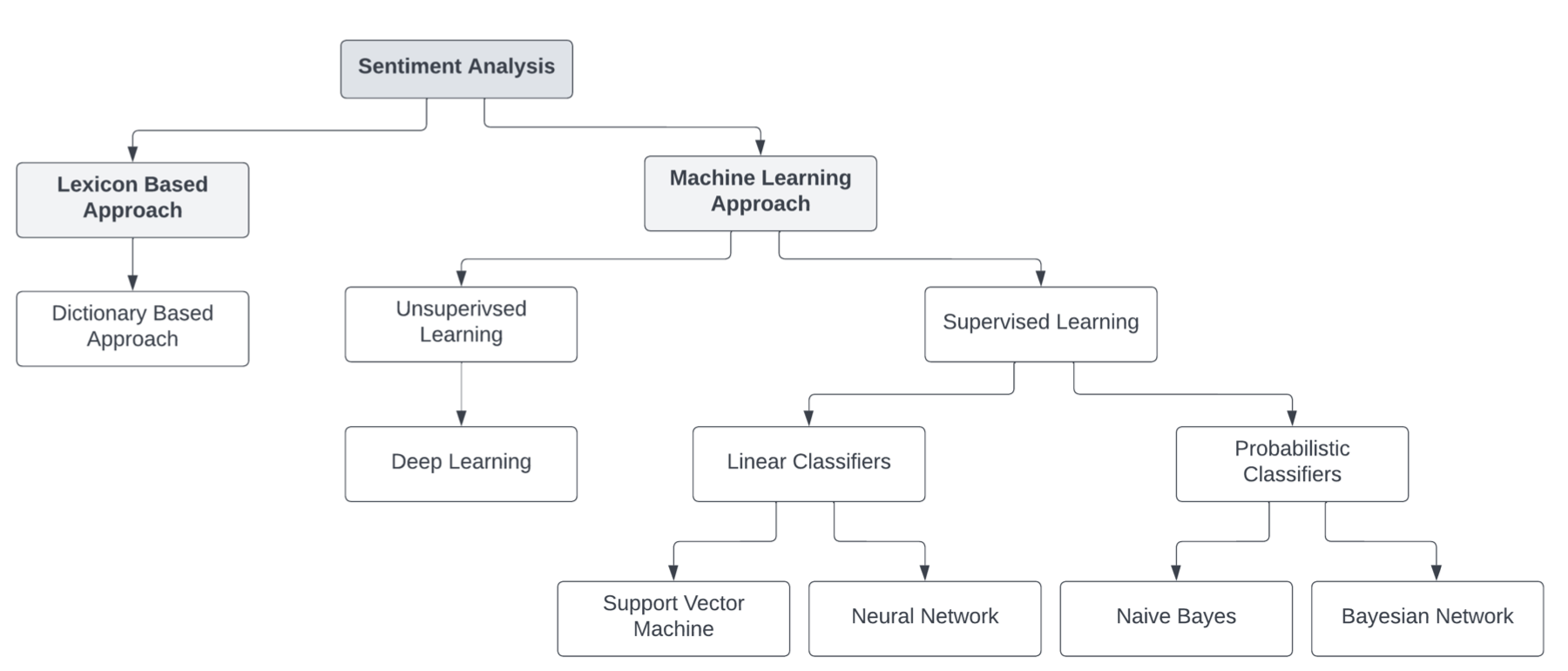}
    \caption{Overview of sentiment analysis methods.}
    \label{sentiment_methods}
\end{figure*}

The potential of sentiment analysis in finance was first recognised  by Fama who, in 1970, introduced the notion of the Efficient Market Hypothesis (EMH) \cite{eff_mark_hypothesis}, which states that stock prices change in response to unexpected fundamental information and news. In this context, before the introduction of advanced machine learning tools, the financial sector employed lexicon-driven approaches \cite{lexicon_approaches}. These methods analyse textual content, sourced from news articles or financial disclosures, based on specific keywords, which are then linked to established sentiment ratings \cite{li2014news,ke2019predicting}. However, an exponential increase in the volume and richness of available information has opened a fertile ground for machine learning, including techniques such as Naive Bayes and Support Vector Machines \cite{cristianini2000introduction}, which are summarised in Figure \ref{sentiment_methods}. \par 
In parallel, the advances in deep learning have become instrumental for NLP research and have spurred  pioneering works that sought to harness the power of neural networks for NLP tasks. Recently, the introduction of the attention mechanism and transformer networks has enabled a significant shift away from  recurrent and convolutional methods, traditionally used in deep-learning tasks \cite{deep_lear_approaches}. This has led to the development of transformer-based models, such as BERT \cite{devlin2019bert}, which owing to its contextual comprehension of language has been used extensively for sentiment analysis. However, the performance of BERT in the financial domain has encountered limitations, primarily because it is not specifically trained on financial datasets. Moreover, its requirement for substantial amounts of data for fine-tuning purposes poses a considerable challenge for financial applications, where such data may not be readily available. \par 
More recently, FinBERT \cite{araci2019finbert}, a version of BERT which is fine-tuned on financial text, has shown promising results for the task of financial sentiment analysis. However, FinBERT still suffers from limitations such as insensitivity to numerical values, while due to its relatively small size (110 million parameters) its classification accuracy deteriorates with  sentence complexity \cite{finbert_complexity}. The FinGPT \cite{liu2023fingpt,yang2023fingpt} and Instruct-FinGPT \cite{zhang2023instructfingpt} aim to enhance their expressive power by using the Llama 7B as their base model. However, FinGPT is not optimized for the task of financial sentiment analysis whilst Instruct-FinGPT only classifies the sentiment valence but is not capable of quantifying the strength of a sentiment class. \par 
To the best of our knowledge, BloombergGPT \cite{wu2023bloomberggpt} is the only pre-trained finance-specific LLM, as Bloomberg was able to train the model using data collected over a span of 40 years. Despite the impressive performance of the model on financial sentiment analysis, the resources required to train such a model are substantial (1.3M GPU hours at a cost of \$5M) whilst much of the training data is confidential and not publicly available. This is different from our proposed methodology, which focuses on achieving a high classification accuracy whilst minimizing the training corpus and computational resources. This is achieved by fine-tuning a pre-trained general-purpose LLM on a smaller-scale financial data corpus. 
    \section{Methodology}
Our work aims to embark upon the immense expressive power and contextual understanding of general-purpose LLMs in order to make them finance-specific. This is achieved by fine-tuning the state-of-the-art (SOTA) Llama 2 7B model on a specific corpus of financial data. The effectiveness of our model is demonstrated on financial sentiment analysis through a new set of benchmarks that align closely with end portfolio construction - the ultimate goal of financial analysis.
\subsection{Fine-tuning the Llama 2 model}
Even though pre-trained LLMs pose a range of capabilities such as reasoning, translation, summarising and text generation, they often struggle when applied to a specific task of interest, such as sentiment analysis. This limitation becomes even more critical in the finance domain, where the nuanced language, media hype and extensive length of financial news articles pose significant additional challenges. \par 
To tackle these challenges, our work revisits the first principles of LLMs in order to align them to the task of financial sentiment analysis. This is achieved by using four labelled financial text datasets as training data to fine-tune the Llama 2 model. Such training on financial data, equipped the model with the ability to understand the linguistic nuances present in the financial domain. Furthermore, a SoftMax classification layer was added at the output of the foundational model, allowing the proposed fine-tuned model to produce SoftMax outputs for three labels: positive, negative or neutral. This made it possible to alter the primary function of the model from text generation to sentiment classification. \par 
\subsubsection{Training datasets}
Our training data was a combination of four labelled publicly available financial news datasets, resulting in a comprehensive collection of 34,180 labelled  samples. Each sample was annotated with one of the three labels: positive, negative, and neutral.
\subsubsection{Model Training}
Our FinLlama model was first initialised with the Llama 2 7B model, followed by fine-tuning over 5 epochs. The training process utilised the AdamW optimizer \cite{Loshchilov2017FixingWD}, as it effectively decouples the weight decay from the optimization steps, leading to more effective training. The initial learning rate was deliberately kept small as the Llama 2 7B model is already pre-trained on a large corpus of data, whilst the warm-up ratio and weight decay served as key regularisation techniques to prevent overfitting, a crucial aspect given the limited size of our fine-tuning dataset. \par 
Moreover, the LoRA implementation was employed in the fine-tuning process in order to minimize the number of trainable parameters whilst achieving high and robust end performance. Through the LoRA implementation, the number of trainable parameters was set to 4.2M, amounting to just 0.0638\% of the total number of parameters in the Llama 2 7B model. This made it possible for our fine-tuning process to be implemented on a \textbf{single} A100 (40 GB) GPU, thus avoiding the need for excessive computational resources. 
\subsection{Proposed Framework}
\begin{figure}[htbp] 
\centering 
\includegraphics[width=0.45\textwidth]{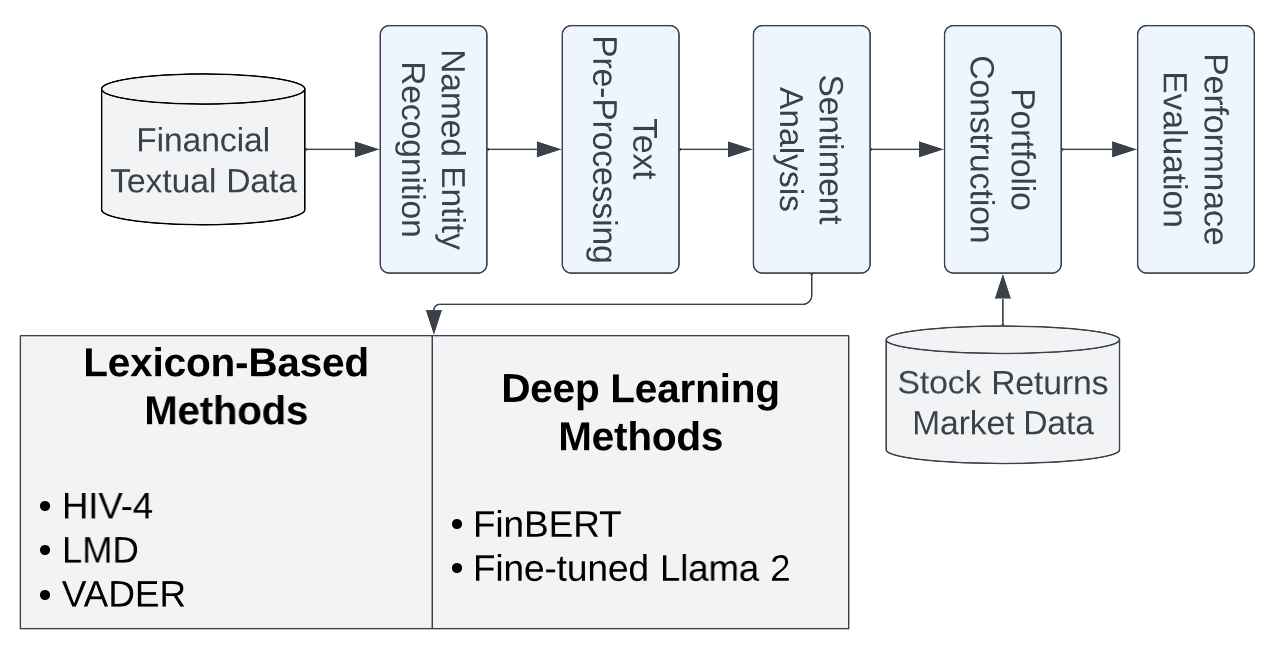}
    \caption{Framework for sentiment analysis.}
    \label{project_framework}
\end{figure}

With the proposed fine-tuned Llama 2 model in place, we followed the framework shown in Figure \ref{project_framework}. Our aim was to assess the performance of our \textit{FinLlama} model against other established sentiment analysis methods, using finance-specific real-world metrics. \par   
\textbf{Data Collection and Processing. }Both textual and market data were collected in order to construct appropriate long-short (L/S) portfolios. Regarding the textual data, 204,017 articles dating between 2015 to 2021 were collected from  online sources. These sources were selected due to their reliability, reputation, lack of bias and focus on major corporations. Financial market data were collected for the same time period from Yahoo Finance. These collected market data contained daily stock returns for the 500 companies in our Investable Universe (S\&P 500), resulting in 1,672 days of stock returns data for each company. Data processing in the form of Named Entity Recognition (NER) and text pre-processing was then applied to the textual data, to remove irrelevant articles and ensure the compatibility of the articles with our sentiment methdods.  \par 
\textbf{Sentiment Analysis.} In total, five sentiment analysis methods were applied. For the lexicon-based approaches (see Appendix \ref{app:lexicon_based}), LMD \cite{lmd} and HIV-4  \cite{hiv4} were implemented using the pysentiment2 Python library, while VADER \cite{vader} was implemented using the NLTK library. Regarding the deep learning methods (see Appendix \ref{app:deep_learning}), both the FinBERT model and our FinLlama model were obtained through HuggingFace, and were utilised via the Transformers library. \par 
The considered methods were evaluated on every article within each corpus for a given company. In cases where multiple articles were published on the same day for a given company, the average sentiment was calculated for that day, in the form
\begin{equation}
\hspace{30mm}S_t=\frac{1}{N_t}\sum_{i=1}^{N_t} S_{it}
\end{equation}
Here, $S_t$ represents the average sentiment for the t-th day, $N_t$ denotes the number of news articles published on that same t-th day for a given company, while $S_{it}$ designates the sentiment strength of the i-th news article on a particular t-th day. The daily sentiment outputs for each company were merged to arrive at the final sentiment data that were utilised as a parameter in the portfolio construction stage. \par 
\textbf{Portfolio Construction.} Once the sentiment for each method was defined for every company, the long-short portfolio was constructed. We used the sentiment as a parameter to determine which companies should be in a long or short position, aiming to maximise returns from both positions. The long-short portfolio was constructed using the following procedure:
\begin{itemize}
    \item \textit{Define the Investable Universe:} Even though the S\&P 500 comprises 500 companies, the financial textual data collected did not contain articles associated to some of the companies for the test period of February 2015 to June 2021. Consequently, 417 companies were considered. 
    \item \textit{Define the long and short position}: The sentiment signal obtained from each of the five methods was used to construct five distinct portfolios. For each method, companies were ranked daily according to their sentiment. Companies that did not have sentiment data on a particular day were omitted from the ranking. As the daily sentiment score for each company ranges between -1 and 1, those with the highest positive sentiment were placed in long positions, whilst those with the strongest negative sentiment were placed in short positions.
    \item \textit{Allocation:} An equally-weighted portfolio strategy was considered in our portfolio construction as this is the strategy mostly utilised by hedge funds \cite{hedge_funds_strat}. The percentage of companies in a long and short position was fixed at 35\%. Consequently, the top 35\% of companies in terms of performance were allocated to long positions, while the bottom 35\% were allocated to short positions.  
    \item \textit{Determine daily returns:} The daily return for each company that was held in a long or short position was obtained by the market data on that particular day. The total daily return of companies that were held in a long position was defined as  
    \begin{equation} 
    Daily \; Long \; Return \;\; r_{long}=\sum_{i=1}^{N_{long}} \frac{r_{long}(i)}{N_{long}}
    \end{equation}
    Similarly, the total daily return of companies that were held in a short position was defined as
    \begin{equation}
    Daily \; Short \; Return \;\; r_{short}=\sum_{i=1}^{N_{short}} \frac{r_{short}(i)}{N_{short}}
    \end{equation}
    For each particular day, the number of companies that were held in either a long position ($N_{long}$) or a short position ($N_{short}$) were equal. Consequently, the total portfolio return on a particular day was the difference between the daily long return and daily short return, and is given by
    \begin{equation}
    Daily \; Return \;\; r_{daily}(i)=r_{long}(i) - r_{short}(i)
    \end{equation}
\end{itemize} \par 
\textbf{Portfolio Evaluation.} The performance of the portfolio constructed using our fine-tuned model was assessed against the portfolios constructed using other SOTA sentiment methods. To this end, the employed real-world financial metrics were: cumulative returns, annualized return, annualized volatility, and  the Sharpe ratio \cite{sharpe_ratio}. These metrics are defined as 
\begin{equation}
    Cumulative \; Returns \;\; r_{cum}=\sum_{i=1}^{N} r_{daily}(i)
\end{equation}
\begin{equation}
    Annualized \; Return \;\; R_p=\sum_{i=1}^N\frac{r_{log}(i)}{N} \times 252
\end{equation}
\begin{equation}
    Ann. \; Volatility \;\; \sigma_p=\sqrt{\frac{\sum_{i=1}^{N}(r_{log}(i)-\bar{r})^2}{N-1}} \times \sqrt{252}
\end{equation}
\vspace{0.2mm}
\begin{equation}
    Sharpe \; Ratio \;\; S_a = \frac{R_p-R_f}{\sigma_p}
\end{equation}
where $N$ is the total number of investing days, totaling 1,672, $r_{log}(i)$ is the logarithmic daily return, $\bar{r}$ is the average daily logarithmic return, $R_f$ is the annualized risk-free rate of return and 252 is the number of business days in a year. The risk-free return, $R_f$, typically represents the yield of the 10-Year Treasury Note; however, due to its prolonged low yield \cite{low_yield} during the analysed period, a 0\% rate is commonly used and is adopted in our analysis. 

    \section{Experimental Results}
The performance of the five portfolios which were constructed as described in Section III are shown in Figure 3. 
\begin{figure*}[htbp] 
\centering 
\includegraphics[width=1\textwidth]{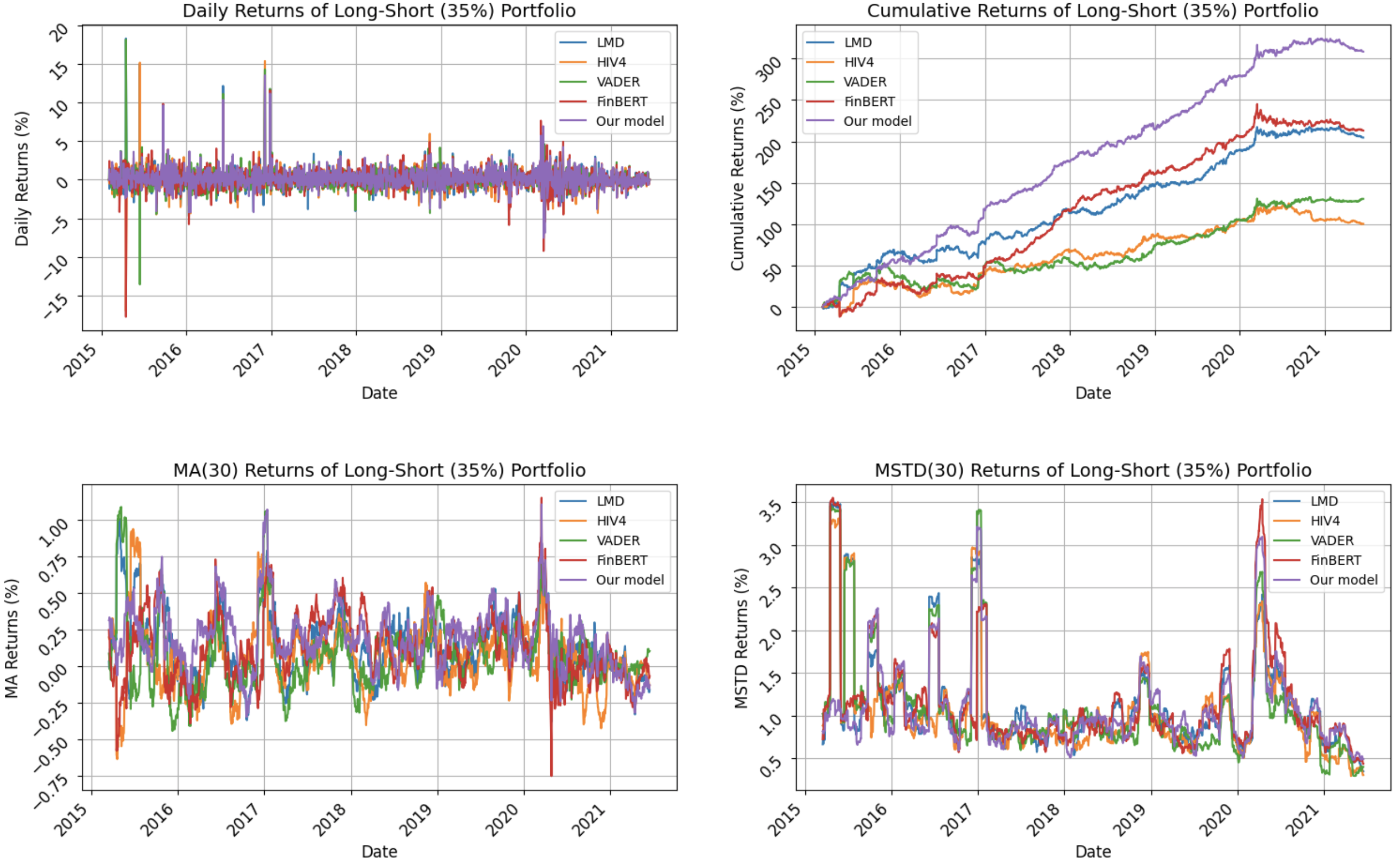}
    \caption{Comparison of performance of the 35\% long-short portfolios which were constructed using the five sentiment analysis methods, for the time period of February 2015 to June 2021. MA(30) and MSTD(30) represent the moving average and moving standard deviation, respectively,  of the returns calculated over a 30-day rolling window.}
    \label{fig:3_3_1_a}
\end{figure*} 
Observe that the deep learning approaches outperformed the lexicon-based approaches in terms of cumulative returns, particularly those relying on general-purpose dictionaries (HIV-4 and VADER). This was to be expected, given that lexicon-based approaches often fail to capture the contextual meaning of sentences, whilst the nuanced nature of financial text significantly reduces the accuracy of general-purpose dictionaries. Furthermore, as observed in the bottom-left panel of Figure 3, all methods exhibited their best performance during turbulent and high-volatility economic periods, such as the first quarter of 2020. The results observed in Table II suggest that the 35\% long-short portfolio, constructed using our fine-tuned Llama-2 model, was the most successful. This is attributed to its ability to achieve significantly higher cumulative returns compared to all other considered methods, and most importantly FinBERT, while attaining a higher Sharpe ratio and exhibiting  lower volatility.  \par 
Overall, our FinLlama model successfully generated significantly higher returns for investors, whilst simultaneously reducing portfolio risk, as indicated by the higher Sharpe ratio and lower annualized volatility.
\begin{table*}[]
\centering
\tiny
\resizebox{\linewidth}{!}{\begin{tabular}{lccccc|c}
\hline
& \textbf{LMD} & \textbf{HIV-4} & \textbf{VADER} & \textbf{FinBERT} & \textbf{FinLlama (Ours)} &\textbf{S\&P 500}  \\ \hline
Cumulative Returns (\%) &   204.6    &   100.4        &      130.6    &     213.0       &  \textbf{308.2}   & 83.1        \\
Annualized Return (\%) &  29.1     &   13.5     &  17.9  &     30.3      &  \textbf{45.0}   & 11.3     \\
Sharpe Ratio      & 1.5        &      0.7      &     0.9    &  1.5            & \textbf{2.4}  & 0.62       \\
Annualized Volatility (\%) & 19.5  & 18.9 & 19.6 & 20.3 & 18.6   & \textbf{18.5} \\
\hline
\end{tabular}}
\caption{Statistical comparison between the five considered sentiment analysis methods using a 35\% long-short portfolio. For Cumulative Returns, Annualized Return and Sharpe Ratio, higher is better. For Annualized Volatility, lower is better. }
\label{tab:my_label}
\end{table*}

    \section{Conclusion and Future Work}
We have introduced an innovative approach to financial sentiment analysis which rests upon the fine-tuning of a general-purpose LLM. In this way, the proposed method has capitalised on the extensive knowledge base and reasoning abilities inherent to LLMs, whilst shifting their primary objective from text generation to classification tasks. In addition, such an approach has enabled the LLMs to become more attuned to the nuanced language of the finance sector, whilst minimising their resource utilisation and computational demands. \par 
Our fine-tuned Llama2 7B model, termed FinLlama, has been used to construct a portfolio, yielding results that have surpassed those of the leading current method in the field, the FinBERT. The FinLlama has achieved cumulative returns which have outperformed the FinBERT model by 44.7\%, while achieving a significantly higher Sharpe ratio and lower annualized volatility. This represents both a substantial contribution to the field of a conjoint framework involving sentiment analysis and LLMs, and demonstrates that fine-tuning an LLM can yield superior results, even with a small amount of task-specific data. In addition, the present work has set a new benchmark in the field, transcending traditional measures such as accuracy and F1-score, commonly used in the literature. Instead, we focus on practical, finance-specific metrics which have greater relevance to end-users. It is our hope that such an approach is a step towards narrowing down the divide between academic research and practical applications within quantitative finance. \par 
Our future research will aim to enhance both the sentiment classification accuracy and efficiency of our model by incorporating additional techniques to produce an easily tractable platform to facilitate the application of artificial intelligence (AI) in the finance sector. \par
\textbf{Disclaimer: Nothing herein is financial advice, and NOT a recommendation to trade real money. Please use common sense and always first consult a professional before trading or investing.}
    \bibliographystyle{IEEEtran}
    \bibliography{references}
    \section{Appendix}
\subsection{Lexicon-Based Approaches}
\subsubsection{Harvard IV-4 Psychological Dictionary (HIV-4)} \label{app:lexicon_based}
HIV-4 is one of the oldest manually constructed lexicons, and is used for objectively identifying specified characteristics of messages in areas involving social science, political science, and psychology. The latest version of the HIV-4 dictionary contains over 11,000 words which are classified into one or more of 183 categories. In this work, we focus on the 1,045 words labelled as positive and the 1,160 words labelled as negative. \newline 
\subsubsection{Loughran and McDonald (LMD) Dictionary}
Loughran and McDonald evaluated standard dictionaries and found that these frequently misclassify terms within financial texts. This insight led to the development of the LMD dictionary, which is specifically tailored for the financial sector. The dictionary categorizes words into six distinct sentiment categories: negative, positive, uncertainty, litigious, strong modal, and weak modal. It was constructed using data from 50,115 10-K filings from 8,341 firms listed on the New York Stock Exchange (NYSE) and the National Association of Securities Dealers Automated Quotations (NASDAQ), covering the period from 1994 to 2008. Overall, the LMD dictionary contains 2,355 negative financial words and 353 positive financial words. \newline
\subsubsection{Valence Aware Dictionary for sEntiment Reasoning (VADER)}
VADER combines lexical features, derived from micro-blog contexts, with the grammatical and syntactical conventions that humans typically employ to express or emphasize sentiment intensity. This enables VADER to accurately quantify the sentiment strength of text. The model contains approximately 9,000 token features, which are each assigned a sentiment score ranging from -4 (indicating extremely negative sentiment) to +4 (indicating extremely positive sentiment). The overall polarity score for a text is calculated by summing the sentiment scores of each word present in the lexicon, with the final score normalized to fall within the range of -1 to +1.

\subsection{Deep Learning Approaches} \label{app:deep_learning}
\subsubsection{FinBERT}
FinBERT leverages the BERT model architecture, and is specifically tailored for financial contexts. It was pre-trained on a substantial financial text corpus consisting of 1.8M news articles sourced from the Thomson Reuters Text Research Collection (TRC2) dataset, spanning the years between 2008 to 2010. Further refinement was achieved through fine-tuning on the Financial Phrasebank (FPB) dataset, thus enhancing its capabilities in financial sentiment classification. FinBERT generates SoftMax outputs for three labels: positive, negative, and neutral.

\newpage

\end{document}